\documentclass[letterpaper]{article} 
\usepackage{aaai2026}  

\usepackage{times}  
\usepackage{helvet}  
\usepackage{courier}  
\usepackage[hyphens]{url}  
\usepackage{graphicx} 
\usepackage{natbib}  
\usepackage{caption} 
\usepackage{booktabs}
\usepackage{siunitx}
\usepackage{multirow}
\usepackage{amssymb}

\usepackage[breakable]{tcolorbox}

\usepackage[utf8]{inputenc}
\usepackage{amsmath}
\usepackage[ruled,vlined,linesnumbered]{algorithm2e}

\usepackage{newfloat}
\usepackage{listings}
\DeclareCaptionStyle{ruled}{labelfont=normalfont,labelsep=colon,strut=off} 

\nocopyright

\title{Promoting Efficient Reasoning with Verifiable Stepwise Reward}
\author{
    Chuhuai Yue\textsuperscript{\rm 1,3},
    Chengqi Dong\textsuperscript{\rm 1,4},
    Yinan Gao\textsuperscript{\rm 2},\\
    Hang He\textsuperscript{\rm 1,5},
    Jiajun Chai\textsuperscript{\rm 1},
    Guojun Yin\textsuperscript{\rm 1}\thanks{Corresponding author},
    Wei Lin\textsuperscript{\rm 1},
}
\affiliations{
    \textsuperscript{\rm 1}Meituan,
    \textsuperscript{\rm 2}Fudan University,
    \textsuperscript{\rm 3}Beijing Institute of Technology\\
    \textsuperscript{\rm 4}University of Science and Technology of China,
    \textsuperscript{\rm 5}East China Normal University\\

}

\usepackage{bibentry}

\begin{document}

\maketitle

\begin{abstract}

Large reasoning models (LRMs) have recently achieved significant progress in complex reasoning tasks, aided by reinforcement learning with verifiable rewards. However, LRMs often suffer from overthinking, expending excessive computation on simple problems and reducing efficiency. Existing efficient reasoning methods typically require accurate task assessment to preset token budgets or select reasoning modes, which limits their flexibility and reliability. In this work, we revisit the essence of overthinking and identify that encouraging effective steps while penalizing ineffective ones is key to its solution. To this end, we propose a novel rule-based verifiable stepwise reward mechanism (VSRM), which assigns rewards based on the performance of intermediate states in the reasoning trajectory. This approach is intuitive and naturally fits the step-by-step nature of reasoning tasks. We conduct extensive experiments on standard mathematical reasoning benchmarks, including AIME24 and AIME25, by integrating VSRM with PPO and Reinforce++. Results show that our method achieves substantial output length reduction while maintaining original reasoning performance, striking an optimal balance between efficiency and accuracy. Further analysis of overthinking frequency and pass@k score before and after training demonstrates that our approach in deed effectively suppresses ineffective steps and encourages effective reasoning, fundamentally alleviating the overthinking problem. All code will be released upon acceptance.

\end{abstract}

\section{Introduction}

Large language models (LLMs) have rapidly emerged as powerful tools for tackling a wide range of general-purpose problems, achieving unprecedented performance in natural language processing and beyond. However, traditional LLMs struggle in tasks that require complex reasoning. Large reasoning models (LRMs), such as O1\cite{openai_o1} and DeepSeek-R1\cite{deepseekai2025deepseekr1incentivizingreasoningcapability}, utilize test-time scaling (TTS) techniques to allocate more computational resources during inference to generate longer Chains-of-Thought (CoTs), significantly enhancing their ability to decompose complex problems, solve them step-by-step, and explore multiple solutions. These advancements have led to substantial breakthroughs in coding and mathematical tasks.

Nevertheless, LRMs are still far from being practically deployable in real-world applications due to their inability to adapt computational depth based on problem complexity. This often leads to the phenomenon of overthinking: allocating excessive computational resources to simple problems, leading to longer inference times, and sometimes even failing to ensure accuracy. This flaw stems from the training mechanism of reinforcement learning with verifiable rewards (RLVR)---for instance, the binary reward function used in R1 encourages the model to over-generate content in the pursuit of correctness, fostering a preference for \textit{being overly redundant rather than potentially wrong}.

\begin{figure}
    \centering
    \includegraphics[width=1\linewidth]{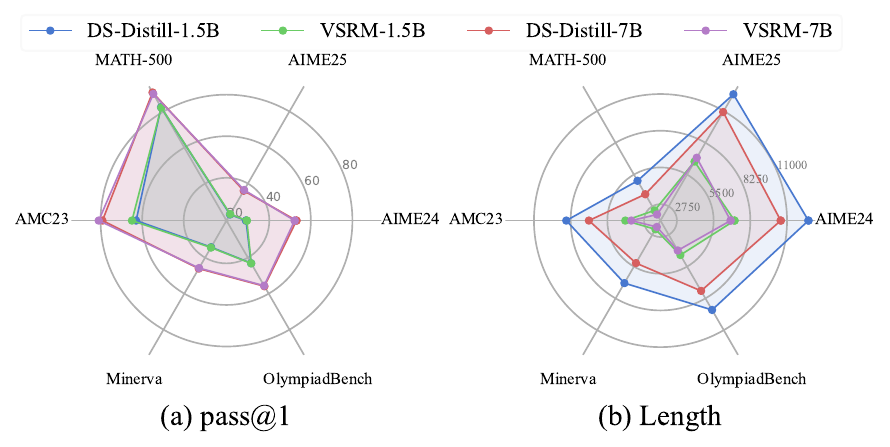}
    \caption{Proposed method substantially reduces token consumption across multiple benchmarks while maintaining model performance.}
    \label{fig:1}
\end{figure}

Existing efforts to optimize LRMs for efficient reasoning generally fall into two main categories. The first category involves setting output length constraints for tasks, essentially imposing a preset token budget.
The second approach, known as difficulty-adaptive reasoning, allocates reasoning modes based on the complexity of the problem.
While this latter strategy appears more logical, it still inherently involves implicit token budgeting. Both methods necessitate pre-evaluating tasks to determine output targets, which not only depends heavily on the accuracy of problem anticipation but also suffers from a lack of flexibility when addressing dynamic scenarios.

Rethinking the essence of the overthinking phenomenon,~\citet{chen2025do} conduct extensive analysis and identify that a key manifestation of overthinking is the model spending substantial computation on intermediate steps that contribute little to final accuracy. Given that complex reasoning tasks inherently unfold in a step-by-step manner, we argue that the fundamental solution to overthinking lies in: encouraging intermediate steps that improve accuracy while penalizing the meaningless ones.

Building on the above insights, we propose a verifiable stepwise reward mechanism (VSRM) to address the overthinking problem in LRMs and enable more efficient reasoning. To provide timely reward signals throughout the reasoning trajectory, we discard the widely used result-based reward model in RLVR settings and instead adopt a stepwise reward framework. Traditional process-level reinforcement learning typically relies on training a process reward model (PRM). However, despite prior successes, PRMs often suffer from instability and limited accuracy. 

To circumvent these issues, our VSRM design eliminates the need for PRMs by leveraging rule-based reward calculation, offering greater interpretability. Specifically, we segment the full reasoning trajectory generated during rollout (referred to as main rollout) into multiple sub-trajectories (referred to as sub-rollouts) of varying lengths , using special tokens that indicate reasoning steps as segmentation points. Each sub-rollout is treated as a prompt to generate multiple responses. Our proposed VSRM then assign stepwise rewards at each truncation point based on the variation in correctness of sub-rollout responses, thereby encouraging/penalizing specific steps. We evaluate our approach primarily on mathematical problems using extensive benchmarks and multiple base models. Experimental results in Figure \ref{fig:1} demonstrate that our method, besides its high interpretability, effectively addresses the essence of the overthinking problem, significantly compressing output length while maintaining or even surpassing the performance of base models. In summary, our contributions can be outlined as follows:

\begin{itemize}
    \item We re-think the essence of the overthinking phenomenon and identified the root solution: encouraging beneficial intermediate steps while penalizing ineffective ones.
    \item We propose VSRM, which assigns verifiable stepwise rewards to the model in a rule-based manner, implementing encouragement and penalization of specific steps, fundamentally optimizing the model.
    \item Our extensive experiments on multiple diverse base models show that our method not only possesses high interpretability but also achieves significant output length compression while maintaining performance.
\end{itemize}

\section{Related Work}

\subsection{Stimulating Reasoning Ability}


The introduction of O1 marks a major advance in reasoning performance and the beginning of the LRM era, inspiring efforts to replicate such strong reasoning abilities. DeepSeek-R1, for example, achieves comparable results using a simple rule-based reward with the group relative policy optimization (GRPO)\cite{shao2024deepseekmathpushinglimitsmathematical} algorithm, and its open-source release has established RLVR as an effective paradigm for improving LLM reasoning. Subsequent models, including the Kimi K series\cite{kimiteam2025kimik15scalingreinforcement,kimik2}, QwQ\cite{QWQ}, and O3\cite{openai_o3}, further advance these capabilities. RLVR assigns scores to trajectories based on pre-designed rules, rewarding desirable behaviors and penalizing undesirable ones. This encourages models to generate long CoTs to maximize correctness, fostering advanced reasoning behaviors such as search and backtracking. However, this also leads to a preference for redundancy over risk of error, resulting in overthinking—wasting computational resources and potentially impairing performance, thus limiting the Practical applicability of LRMs.

\subsection{Attempts Towards Efficient Reasoning}
\label{2.2}
Overthinking issue is first identified and analyzed by \citet{chen2025do}, who observe that LRMs generate lengthy outputs that neither improve accuracy nor introduce new solution strategies especially for easy prompt. To address this, various works explore efficient reasoning from different angles. \citet{huang2025adacotrethinkingcrosslingualfactual}, \citet{xu2025chaindraftthinkingfaster}, and \citet{han2025tokenbudgetawarellmreasoning} focus on prompt engineering to encourage concise responses, while \citet{yang2025dynamicearlyexitreasoning} and \citet{fan2025cyclicrefleximprovinglargereasoning} target decoding strategies to suppress unnecessary CoT extensions. Although these training-free approaches can alleviate overthinking, they cannot fundamentally eliminate the problem. \citet{ma2025cotvalvelengthcompressiblechainofthoughttuning}, \citet{ye2025limoreasoning}, and \citet{yu2025longshortchainofthoughtmixturesupervised} employ supervised fine-tuning (SFT) on well-curated datasets to help models memorize efficient reasoning patterns, while \citet{zhang2025lightthinkerthinkingstepbystepcompression} and \citet{luo2025o1prunerlengthharmonizingfinetuningo1like} explicitly incorporate efficiency as an optimization objective.

A more mainstream approach is to use reinforcement learning for better generalization. \citet{kimiteam2025kimik15scalingreinforcement}, \citet{shen2025dastdifficultyadaptiveslowthinkinglarge}, \citet{yeo2025demystifying}, and \citet{cheng2025optimizinglengthcompressionlarge} introduce length penalties in the reward function to suppress overly long reasoning traces. \citet{arora2025traininglanguagemodelsreason} and \citet{zhang2025grpoleaddifficultyawarereinforcementlearning} use intra-group length regularization to encourage shorter correct answers. \citet{yi2025shorterbetterguidingreasoningmodels}, \citet{hou2025thinkprunepruninglongchainofthought}, and \citet{qi2025optimizinganytimereasoningbudget} optimize performance under a fixed token budget to balance efficiency and effectiveness. Approaches such as \citet{zhang2025adaptthinkreasoningmodelslearn}, \citet{huang2025adactrladaptivecontrollablereasoning}, and \citet{wu2025armadaptivereasoningmodel} assign predefined thinking patterns based on task difficulty, which essentially reflects a length budget. All the aforementioned methods heavily rely on presetting a rough length estimation for different problems and optimize the model by comparing the gap between this estimation and the actual rollout. Such estimation lacks both reliability and flexibility. Moreover, whether explicitly or implicitly, incorporating length compression as part of the optimization objective may potentially impair the model's reasoning capability.

Our work revisits the essence of overthinking and, leveraging the sequential nature of reasoning tasks, encourages effective steps while penalizing ineffective ones through stepwise rewards. The proposed method is highly interpretable and promotes efficient reasoning without compromising performance.

\section{Methodology}

\subsection{Rethink the Essence of Overthinking Issue}
\label{sec:essence}
Based on the analysis in the previous section, in order to enable LRMs to perform efficient reasoning elegantly, we first need to rethink the essence of the overthinking issue.

\citet{chen2025do} use QwQ and R1 to generate responses for multiple evaluation benchmarks and observe a general phenomenon of “one problem, multiple solutions,” especially for simpler questions. To gain a finer-grained understanding of overthinking and more precisely locate its root cause, we build upon this observation and take a step further.

Reasoning tasks—particularly in mathematical reasoning and code generation—typically require step-by-step answers. Leveraging this, we use DeepSeek-distilled-Qwen-1.5B\cite{deepseekai2025deepseekr1incentivizingreasoningcapability} and DeepScaleR-1.5B\cite{deepscaler2025} (referred to as DS-Distill-1.5B and DeepScaleR, respectively) to generate answers for MATH-500\cite{lightman2023letsverifystepstep} problems. From the 500 generated responses, we randomly sample 10\% and annotate the reasoning steps for a case study. We find that the reasoning process of LRMs can generally be divided into three stages: problem restatement, step-by-step answering, and answer summary. The overthinking issue mainly occurs during the step-by-step answering stage. As shown in Figure~\ref{fig:case}, different steps are highlighted in different colors. In the example, the response is excessively verbose, expending substantial computational resources on a simple question—how many integers less than $0$ are there in [-500,0]—and ultimately arriving at an incorrect conclusion. Many clearly ineffective steps neither improve accuracy nor help the model reach the correct answer. Such cases are not isolated in our study.

\begin{figure}
    \centering
    \includegraphics[width=0.9\linewidth]{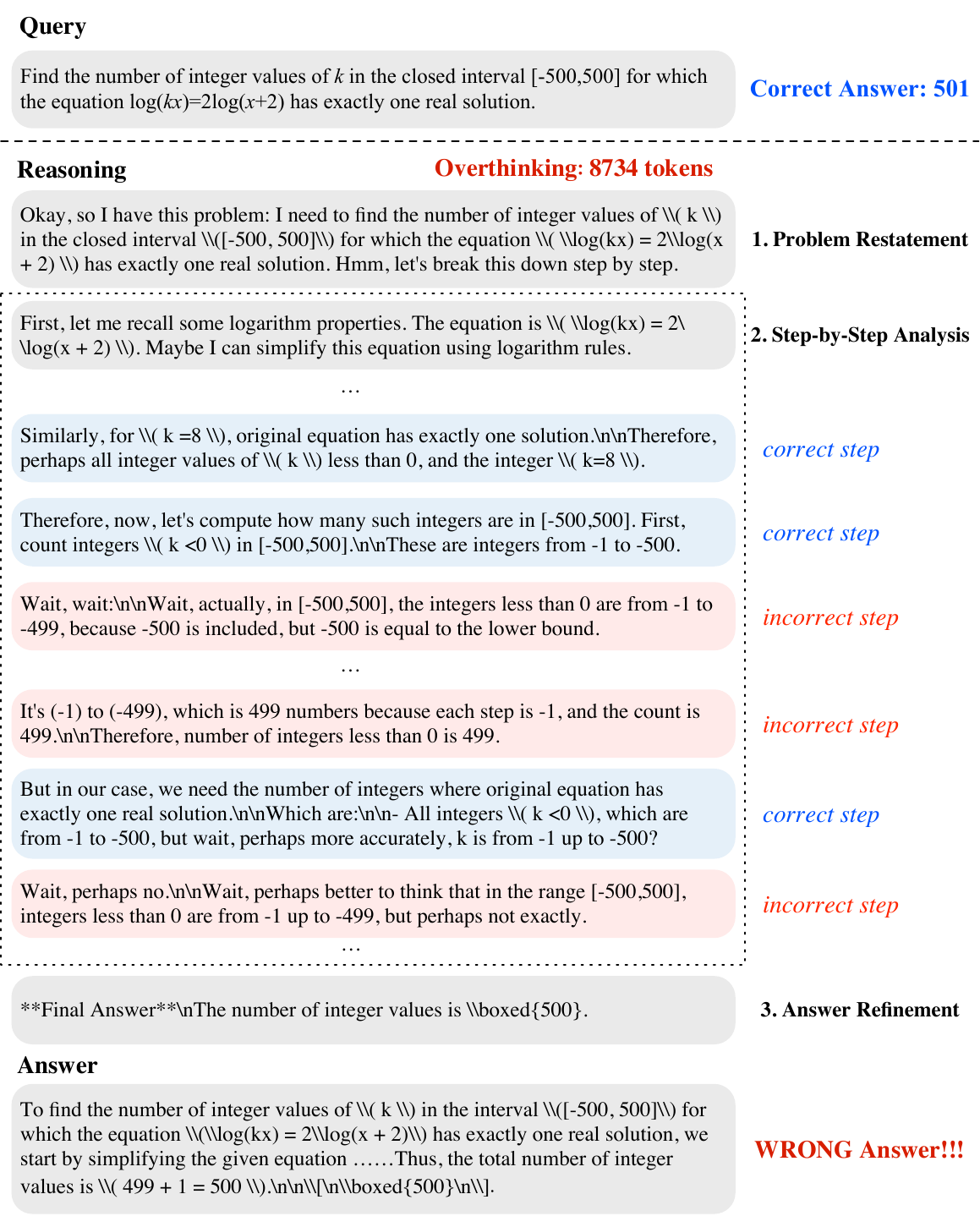}
    \caption{Illustration of the step-by-step answering process by DeepScaleR, with \textit{correct} and \textit{incorrect} steps highlighted in different colors.}
    \label{fig:case}
\end{figure}

\begin{table}[h!]
\centering
\footnotesize
\begin{tabular}{l|c}
\toprule
\textbf{Model} & \textbf{Detected Overthinking} \\
\midrule
DeepScaleR & 279/500 \\
DS-Distill-1.5B & 312/500 \\
\bottomrule
\end{tabular}
\caption{Detecting ineffective steps via DeepSeek-R1}
\label{tb1}
\end{table}

To obtain a more comprehensive view, we further prompt DeepSeek-R1 to analyze all 500 responses generated by both models and statistically examine the occurrence of ineffective steps. The prompt used is provided in the Appendix, and results are summarized in Table~\ref{tb1}. Comparing DeepSeek-R1's predictions with our 50 manual annotations, we find that although it does not identify all ineffective steps precisely, it is able to detect overthinking phenomena, making the statistics generally trustworthy.

Based on the analysis of R1, both models exhibit significant ineffective steps in over half of the samples. This indicates that, in general scenarios, the generation of numerous ineffective steps by the models constitutes the primary manifestation of the overthinking phenomenon. Therefore, we propose that the key to alleviating overthinking is to accurately distinguish between effective and ineffective steps, and to set the optimization objective as encouraging effective steps and penalizing ineffective ones, thereby achieving efficient reasoning. Compared to directly compressing the length of the CoT, this is more intuitive and is closely aligned with the step-by-step nature of reasoning tasks.

\begin{figure*}
    \centering
    \includegraphics[width=0.80\linewidth]{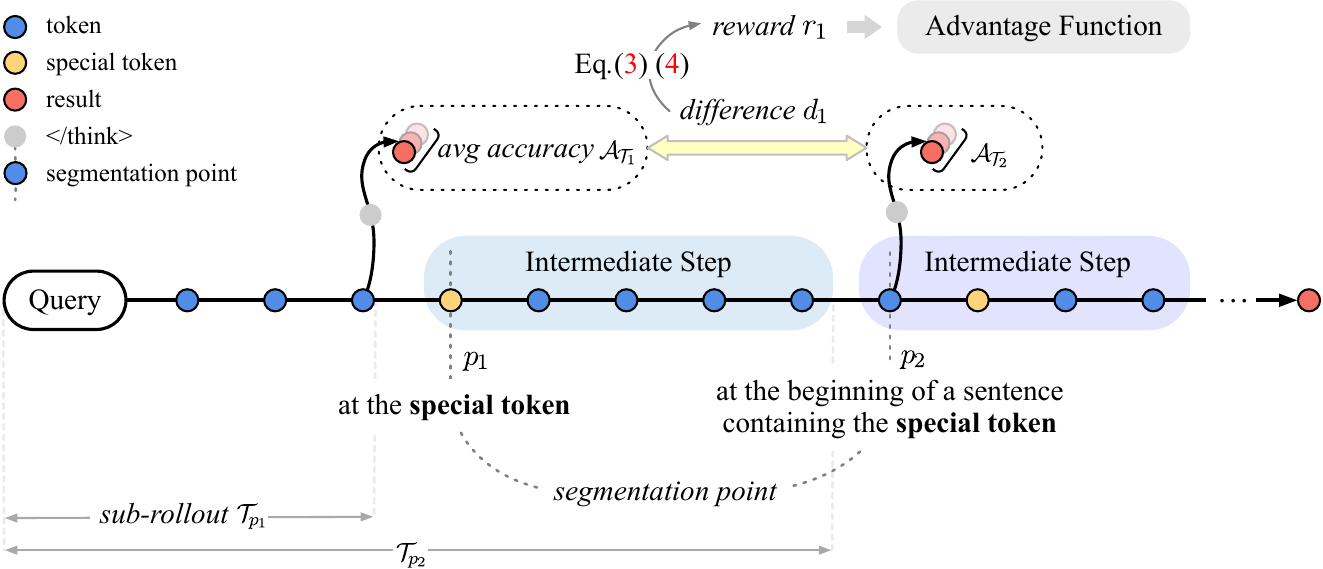}
    \caption{For a given main rollout $\mathcal{T}$, the process of obtaining its complete reward information involves the following steps: segmenting $\mathcal{T}$ into sub-rollouts; generating multiple candidate answers for each sub-rollout; assigning stepwise rewards based on the average accuracy of all sub-rollouts via proposed method; and finally, combining these with the outcome reward to compute the advantage.}
    \label{fig:enter-label}
\end{figure*}

\subsection{Make Stepwise Reward Verifiable}

The most commonly used outcome-based reward mechanism in RLVR is incompatible with our optimization objective mentioned above as it cannot provide timely intermediate rewards at the reasoning process’s intermediate states to reward or penalize different steps. To address this, we propose to apply a stepwise reward mechanism, which naturally aligns with our optimization objective.

A common approach to implementing stepwise reward is to train a PRM that decomposes the reasoning CoT and assigns a score to each stage, guiding the model to generate more interpretable and rigorous reasoning processes\cite{lightman2023letsverifystepstep}. This approach has achieved promising results in tasks like mathematical reasoning. However, PRMs suffer from significant limitations in large-scale reinforcement learning scenarios: they are difficult to train and their reliability cannot be guaranteed\cite{deepseekai2025deepseekr1incentivizingreasoningcapability}. This dilemma motivates us to design a verifiable stepwise reward mechanism, which combines the advantages of stepwise rewards with the reliability and usability of verifiable outcome-based rewards. In this way, we circumvent the shortcomings of PRMs while providing interpretable stepwise rewards for each reasoning step.

\subsubsection{Step Seperation.}
The first phase of implementing VSRM is the seperation of reasoning steps. \citet{chen2025do} and \citet{cheng2025optimizinglengthcompressionlarge} attempt to prompt or train an additional model specifically to identify different solutions or correct answers in the chain-of-thought (CoT). However, introducing another model inevitably brings additional uncertainty. Therefore, we design a comprehensive rule-based step segmentation algorithm.

Through the analysis of the generated responses mentioned above, we observe that the boundaries between reasoning steps are often accompanied by certain special tokens, such as “however”, “thus”, “so”, “but” “wait”, etc. These tokens indicate that the model has completed one reasoning step and is about to proceed to the next, whether it is a progression or a transition. \cite{wang20258020rulehighentropyminority} also reported similar phenomena in their work and conducted detailed analysis, which verifies the correctness of our observation. This demonstrates that extracting special tokens from the CoT and using them for step segmentation is feasible.

Specifically, for a complete rollout generated by reference model during reinforcement learning, which will be denoted as main rollout $\mathcal{T}$ in the following, we first use regular expressions to extract the reasoning part (the content within \texttt{\textless think\textgreater}...\texttt{\textless /think\textgreater}). After obtaining the full thinking content, we skip a certain number of initial tokens to avoid splitting the problem restatement part. During segmentation, we use a predefined list of special tokens to divide $\mathcal{T}$. To ensure readability, we introduce two additional rules: (1) there must be at least a minimum distance between adjacent segmentation points, and (2) each segmentation point should align with the beginning of a complete sentence containing the special token. These rules help avoid overly dense segmentation when multiple special tokens appear in a single sentence, and ensure that each resulting step contains complete semantic information. The entire step segmentation procedure, as illustrated in Figure~\ref{fig:enter-label}, results in a set of segmentation points $\mathcal{P} = \{p_1, p_2, \dots, p_K\}$. Further details can be found in the corresponding pseudocode in Appendix.

\subsubsection{Assigning Reward to Intermediate Stage.}

Based on $\mathcal{P}$, we can extract the reasoning content from the start up to each $p_i$ and combine it with the input question. In this way, a set of sub-rollouts $\mathcal{T_\mathcal{P}} = \{\mathcal{T}_{p_1}, \mathcal{T}_{p_2}, \dots, \mathcal{T}_{p_k}\}$. are derived from $\mathcal{T}$, each representing a distinct segment of the overall reasoning process. Thanks to the rules established during step segmentation, each $\mathcal{T}_{p_i}$ retains complete semantic information and does not contain incomplete sentences. We append a \texttt{\textless /think\textgreater} token to every $\mathcal{T}_{p_i}$ to mark the end of the reasoning and then use them as new queries for the model to generate corresponding answers. In this way, the reward at position $p_i$ in $\mathcal{T}$ can be directly evaluated by the outcome of $\mathcal{T}_{p_i}$. This effectively transforms the challenge of assigning stepwise rewards to intermediate states into a verifiable and explainable procedure, where the reward for each sub-rollout' result serves as a proxy for the reward of the intermediate state of the main rollout.

The simplest approach to assigning reward signals to each $\mathcal{T}_{p_i}$ is to adopt the commonly used binary reward mechanism, where a correct answer receives a reward of $1$ and an incorrect answer receives $0$. To ensure the stability of the reward signal and reduce the high variance caused by randomness, we generate multiple candidate answers for each $\mathcal{T}_i$, and use the average accuracy as its reward, denoted as $\mathcal{A}_{\mathcal{T}_{i}} \in \mathcal{A}_{\mathcal{T}}$. For a set of $N$ candidate answers generated from $\mathcal{T}_i$, $\mathcal{A}_{\mathcal{T}_i}$ is empirically estimated as:
\begin{equation}
\label{eq:state_value}
\mathcal{A}_{\mathcal{T}_i} = \frac{1}{N} \sum_{j=i}^{N} \mathbb{I}(\text{IsCorrect}(\text{LRM}(\mathcal{T}_i)_j))
\end{equation}
where $\mathbb{I}(\cdot)$ is the indicator function.

However, it is suboptimal to directly assign $\mathcal{A}_i$ as the stepwise reward for each step, since the reward at every position would always be non-negative, making it ineffective in penalizing ineffective steps. Therefore, we instead use the difference in average accuracy between adjacent steps as the reward for each step, as follows:
\begin{equation}
    d_{i-1} = \mathcal{A}_i - \mathcal{A}_{i-1}, \quad \text{for } i\in[1,  \dots, k]
\end{equation}
For the boundary case $r_{k}$, we compute the difference between the average accuracy of $\mathcal{T}$ (denoted as $\mathcal{A}_{\mathcal{T}}$) and $\mathcal{A}_k$.

Assigning the stepwise reward as the difference in average accuracy between consecutive steps encourages the main rollout to evolve towards better states. In this scheme, steps that contribute to an improvement in accuracy are rewarded, while the rest are penalized or suppressed. However, in real-world reasoning processes, arriving at the correct conclusion often requires the accumulation of several intermediate steps, rather than relying on a single decisive action. Consequently, under this difference-based reward assignment, $\mathcal{A}_i$ tends to remain unchanged across many steps, resulting in $d_{i} = 0$ at most positions. This leads to sparse reward signals, thus hinders the training process.

To address this issue, when the difference in average accuracy between consecutive steps is negligible, i.e., $|d_{i-1}| < \epsilon$ for a small threshold $\epsilon$, we further examine subsequent steps to determine if a significant change in accuracy occurs within a limited lookahead window of $L_{\max}$ steps. Specifically, let $q$ be the smallest integer such that $q \leq L_{\max}$ and $|d_{i+q-1}| \geq \epsilon$. If such a $q$ exists, we assign a discounted reward to step $i-1$ as follows:
\begin{equation}
    r_{i-1} = \mathrm{sgn}(d_{i+q-1}) \cdot |d_{i+q-1}| \cdot \gamma^{q},
\end{equation}
where $\gamma \in (0,1)$ is a discount factor and $\mathrm{sgn}(\cdot)$ is an indicator function. If no significant change is found within $L_{\max}$ steps, we set $r_{i-1} = 0$. For the case where the end of the sequence is reached without observing a significant change, we use the difference between the final accuracy and the current accuracy, also discounted appropriately:
\begin{equation}
    r_{i-1} = \mathrm{sgn}(d_{k}) \cdot |d_{k}| \cdot \gamma^{k-i+1}.
\end{equation}

This decay-based propagation mechanism alleviates the excessive sparsity of reward signals, enabling the model to more rapidly cultivate efficient reasoning behaviors. More details about the reward design can be found in Appendix.

\subsection{Reinforcement Learning with VSRM}

Through our proposed VSRM, we are now able to assign stepwise rewards to each main trajectory generated during RL training. We combine the traditional binary outcome reward, the format reward, and stepwise reward to ensure that the LRM's outputs conform to verifiable standards, avoid uncontrolled content generation, and fundamentally guarantee result correctness. Ultimately, we obtain a reward list whose length matches the total response length, which is used for subsequent advantage estimation. The overall reward can be expressed as:
\begin{equation}
    R_{\mathcal{T}} = [\dots,r_1, \dots, r_t, \dots,r_k,\dots,r_{\mathcal{T}}^{\mathrm{result}} + r_{\mathcal{T}}^{\mathrm{format}}] 
\end{equation}
where $r_t$ denotes the stepwise reward, $r_t^{\mathrm{result}}$ is the outcome reward (assigned only at the final position) and $r_t^{\mathrm{format}}$ is the format reward (also assigned only at the final position). For other positions, reward remains zero.

Our proposed VSRM is fully compatible with RL algorithms that support stepwise rewards, such as PPO\cite{schulman2017proximalpolicyoptimizationalgorithms} and Reinforce++\cite{hu2025reinforce++}, and can be seamlessly integrated. In the next section, we will discuss the effectiveness of VSRM through experiments.

\section{Experiments}

\subsection{Experimental Setup}

To thoroughly evaluate the effectiveness of our approach, we conduct experiments using several widely adopted LRMs as base models, including 
DS-Distill-1.5B, DS-Distill-7B\cite{deepseekai2025deepseekr1incentivizingreasoningcapability}, and DeepScaleR\cite{deepscaler2025}. We employ the VeRL\cite{sheng2024hybridflow} framework and apply both PPO and Reinforce++ algorithms, which are naturally compatible with stepwise reward, for post-training of the base models.

For training data, we use the dataset introduced by DeepScaleR, which is compiled from historical AIME problems (1984–2023), historical AMC problems (before 2023), the Omni-MATH dataset, and the Still dataset. This collection covers a wide range of difficulty levels.

In terms of evaluation, we follow previous work and select a comprehensive set of benchmarks, including MATH-500, AIME24, AIME25\cite{AoPS_AIME}, AMC23\cite{MAA_AMC}, Minerva\cite{lewkowycz2022solving}, and OlympiadBench\cite{he-etal-2024-olympiadbench}, providing broader coverage than previous studies. During evaluation, we set the maximum generation length at 20,000 tokens—substantially higher than the maximum response length during training—to ensure that model performance is not constrained by this hyperparameter. The temperature and top-p are set to 0.6 and 0.95, respectively, consistent with \citet{hou2025thinkprunepruninglongchainofthought}'s work and DeepSeek-R1. For all benchmarks, we report the average pass@1 over N runs. Specifically, for MATH-500 and OlympiadBench, where the benchmark sizes are relatively large, we set N to 16; for the other benchmarks, we set N to 32 to reduce randomness.

Additional details can be found in Appendix.

\subsection{Main Results}

\begin{table*}[t]
\centering
\small
\resizebox{\textwidth}{!}{%
\renewcommand{\arraystretch}{1.2}
\begin{tabular}{lcccccccccccc}
\toprule
\multirow{2.5}{*}{\textbf{Method}} & \multicolumn{2}{c}{\textbf{AIME24}} & \multicolumn{2}{c}{\textbf{AIME25}} & \multicolumn{2}{c}{\textbf{MATH-500}} & \multicolumn{2}{c}{\textbf{AMC23}} & \multicolumn{2}{c}{\textbf{Minerva}} & \multicolumn{2}{c}{\textbf{OlympiadBench}} \\

\cmidrule(lr){2-3} \cmidrule(lr){4-5} \cmidrule(lr){6-7} \cmidrule(lr){8-9} \cmidrule(lr){10-11} \cmidrule(l){12-13}

& pass@1$\uparrow$ & Length$\downarrow$ & pass@1$\uparrow$ & Length$\downarrow$ & pass@1$\uparrow$ & Length$\downarrow$ & pass@1$\uparrow$ & Length$\downarrow$ & pass@1$\uparrow$ & Length$\downarrow$ & pass@1$\uparrow$ & Length$\downarrow$ \\
\midrule
\textbf{DS-Distill-1.5B} & 28.8 & 12605 & 22.8 & 12444 & 82.4 & 4960 & 62.9 & 8577 & 34.2 & 6929 & 43.1 & 9249 \\
+ LC-R1 & 26.3 {\scriptsize($\downarrow$2.5)} & 7098 {\scriptsize($\downarrow$5507)} & 20.9 {\scriptsize($\downarrow$1.9)} & 6942 {\scriptsize($\downarrow$5502)} & 80.4 {\scriptsize($\downarrow$2.0)} & 2473 {\scriptsize($\downarrow$2487)} & 62.8 {\scriptsize($\downarrow$0.1)} & 4389 {\scriptsize($\downarrow$4188)} & 32.8 {\scriptsize($\downarrow$1.4)} & 2512 {\scriptsize($\downarrow$4417)} & 42.0 {\scriptsize($\downarrow$1.1)} & 4632 {\scriptsize($\downarrow$4617)} \\
+ ThinkPrune-2k & 26.7 {\scriptsize($\downarrow$2.1)} & 7085 {\scriptsize($\downarrow$5520)} & 19.7 {\scriptsize($\downarrow$2.1)} & 6918 {\scriptsize($\downarrow$5526)} & 81.7 {\scriptsize($\downarrow$0.7)} & 2426 {\scriptsize($\downarrow$2534)} & 63.8 {\scriptsize($\uparrow$0.9)} & 4224 {\scriptsize($\downarrow$4353)} & 33.9 {\scriptsize($\downarrow$0.3)} & 2667 {\scriptsize($\downarrow$4262)} & 42.9 {\scriptsize($\downarrow$0.2)} & 4752 {\scriptsize($\downarrow$4497)} \\
+ AdaptThink & 28.7 {\scriptsize($\downarrow$0.1)} & 8055 {\scriptsize($\downarrow$4550)} & 21.8 {\scriptsize($\downarrow$1.0)} & 8155 {\scriptsize($\downarrow$4289)} & 80.4 {\scriptsize($\downarrow$2.0)} & \textbf{1963} {\scriptsize($\downarrow$2997)} & 63.3 {\scriptsize($\uparrow$0.4)} & \textbf{4069} {\scriptsize($\downarrow$4508)} & 32.3 {\scriptsize($\downarrow$1.9)} & \textbf{1912} {\scriptsize($\downarrow$5017)} & 42.6 {\scriptsize($\downarrow$0.5)} & 4563 {\scriptsize($\downarrow$4686)} \\
+ Efficient Reasoning & \underline{29.2} {\scriptsize($\uparrow$0.4)} & 9189 {\scriptsize($\downarrow$3416)} & \underline{22.9} {\scriptsize($\uparrow$0.1)} & 8590 {\scriptsize($\downarrow$3854)} & \underline{82.0} {\scriptsize($\downarrow$0.4)} & 2621 {\scriptsize($\downarrow$2339)} & \underline{64.7} {\scriptsize($\uparrow$1.8)} & 5202 {\scriptsize($\downarrow$3375)} & \underline{34.4} {\scriptsize($\uparrow$0.2)} & 3230 {\scriptsize($\downarrow$3699)} & \textbf{43.8} {\scriptsize($\uparrow$0.7)} & 5755 {\scriptsize($\downarrow$3494)} \\
+ \textbf{VSRM-PPO (ours)} & \underline{29.2} {\scriptsize($\uparrow$0.4)} & \underline{7065} {\scriptsize($\downarrow$5540)} & \textbf{23.0} {\scriptsize($\uparrow$0.2)} & \textbf{6621} {\scriptsize($\downarrow$5823)} & \textbf{82.2} {\scriptsize($\downarrow$0.2)} & \underline{2400} {\scriptsize($\downarrow$2560)} & \textbf{64.9} {\scriptsize($\uparrow$2.0)} & 4153 {\scriptsize($\downarrow$4424)} & \underline{34.4} {\scriptsize($\uparrow$0.2)} & 2273 {\scriptsize($\downarrow$4656)} & \underline{43.2} {\scriptsize($\uparrow$0.1)} & \textbf{4499} {\scriptsize($\downarrow$4750)} \\
+ \textbf{VSRM-R++ (ours)} & \textbf{29.5} {\scriptsize($\uparrow$0.7)} & \textbf{6958} {\scriptsize($\downarrow$5647)} & \underline{22.9} {\scriptsize($\uparrow$0.1)} & \underline{6892} {\scriptsize($\downarrow$5552)} & 81.7 {\scriptsize($\downarrow$0.7)} & 2597 {\scriptsize($\downarrow$2363)} & \underline{64.7} {\scriptsize($\uparrow$1.8)} & \underline{4119}{\scriptsize($\downarrow$4458)} & \textbf{34.6}
{\scriptsize($\uparrow$0.4)} & \underline{2172} {\scriptsize($\downarrow$4751)} & 43.1 {\scriptsize($\uparrow$0.0)} & \underline{4544} {\scriptsize($\downarrow$4705)} \\
\midrule
\textbf{DS-Distill-7B} & 53.1 & 10529 & 36.0 & 10909 & 90.6 & 3796 & 79.1 & 6884 & 46.1 & 5195 & 55.9 & 7602 \\
+ LC-R1 & 51.7 {\scriptsize($\downarrow$1.4)} & 6820 {\scriptsize($\downarrow$3809)} & 35.7 {\scriptsize($\downarrow$0.3)} & 7458 {\scriptsize($\downarrow$3451)} & 88.3 {\scriptsize($\downarrow$2.3)} & \textbf{1776} {\scriptsize($\downarrow$2020)} & 79.1 {\scriptsize($\uparrow$0.0)} & \textbf{3686} {\scriptsize($\downarrow$3198)} & 44.4 {\scriptsize($\downarrow$1.7)} & \textbf{1834} {\scriptsize($\downarrow$3361)} & 55.5 {\scriptsize($\downarrow$0.4)} & 4193 {\scriptsize($\downarrow$3409)} \\
+ AdaptThink & \underline{52.1} {\scriptsize($\downarrow$1.0)} & 8679 {\scriptsize($\downarrow$1830)} & 35.0 {\scriptsize($\downarrow$1.0)} & 9807 {\scriptsize($\downarrow$1102)} & 88.9 {\scriptsize($\downarrow$1.7)} & 2199 {\scriptsize($\downarrow$1597)} & 80.7 {\scriptsize($\uparrow$1.6)} & 5130 {\scriptsize($\downarrow$1754)} & 45.2 {\scriptsize($\downarrow$0.9)} & 2869 {\scriptsize($\downarrow$2326)} & 55.4 {\scriptsize($\downarrow$0.5)} & 5915 {\scriptsize($\downarrow$1687)} \\
+ Efficient Reasoning & 51.9 {\scriptsize($\downarrow$1.2)} & 8667 {\scriptsize($\downarrow$1862)} & \underline{36.2}
{\scriptsize($\uparrow$0.2)} & 9100 {\scriptsize($\downarrow$1809)} & \textbf{89.8} {\scriptsize($\downarrow$0.8)} & 2408 {\scriptsize($\downarrow$1388)} & 80.7 {\scriptsize($\uparrow$1.6)} & 4933 {\scriptsize($\downarrow$1951)} & 45.7 {\scriptsize($\downarrow$0.4)} & 2903 {\scriptsize($\downarrow$2292)} & \underline{55.6} {\scriptsize($\downarrow$0.3)} & 5599 {\scriptsize($\downarrow$2003)} \\
+ GRPO-LEAD & \underline{52.1} {\scriptsize($\downarrow$1.0)} & 9023 {\scriptsize($\downarrow$1506)} & 36.1 {\scriptsize($\uparrow$0.1)} & 9842 {\scriptsize($\downarrow$1067)} & 89.5 {\scriptsize($\downarrow$1.1)} & 3152 {\scriptsize($\downarrow$644)} & 80.7 {\scriptsize($\uparrow$1.6)} & 5584 {\scriptsize($\downarrow$1300)} & \textbf{46.3} {\scriptsize($\uparrow$0.2)} & 3930 {\scriptsize($\downarrow$1265)} & \textbf{56.1} {\scriptsize($\uparrow$0.2)} & 6372 {\scriptsize($\downarrow$1230)} \\
+ \textbf{VSRM-PPO (ours)} & 51.8 {\scriptsize($\downarrow$1.3)} & \textbf{6675} {\scriptsize($\downarrow$3854)} & \underline{36.2} {\scriptsize($\uparrow$0.2)} & \textbf{6920} {\scriptsize($\downarrow$3989)} & \underline{89.6} {\scriptsize($\downarrow$1.0)} & \underline{2024} {\scriptsize($\downarrow$1772)} & \textbf{81.0} {\scriptsize($\uparrow$1.9)} & 3759 {\scriptsize($\downarrow$3125)} & 45.5 {\scriptsize($\downarrow$0.6)} & \underline{2002} {\scriptsize($\downarrow$3193)} & 55.3 {\scriptsize($\downarrow$0.6)} & \underline{4168} {\scriptsize($\downarrow$3434)} \\
+ \textbf{VSRM-R++ (ours)} & \textbf{52.2} {\scriptsize($\downarrow$0.9)} & \underline{6773} {\scriptsize($\downarrow$3756)} & \textbf{36.4} {\scriptsize($\uparrow$0.4)} & \underline{6953} {\scriptsize($\downarrow$3956)} & \textbf{89.8} {\scriptsize($\downarrow$0.8)} & 2044 {\scriptsize($\downarrow$1752)} & \underline{80.9} {\scriptsize($\uparrow$1.8)} & \underline{3704} {\scriptsize($\downarrow$3180)} & \underline{45.8} {\scriptsize($\downarrow$0.3)} & 2068 {\scriptsize($\downarrow$3127)} & \underline{55.6} {\scriptsize($\downarrow$0.3)} & \textbf{4103} {\scriptsize($\downarrow$3499)} \\
\midrule
\textbf{DeepScaleR} & 40.7 & 8515 & 31.3 & 8029 & 86.5 & 3105 & 73.6 & 5487 & 38.9 & 5102 & 50.0 & 5647 \\
+ ThinkPrune-2k  & 36.5 {\scriptsize($\downarrow$4.2)} & \textbf{6108} {\scriptsize($\downarrow$2407)} & 26.0 {\scriptsize($\downarrow$5.3)} & \textbf{5486} {\scriptsize($\downarrow$2543)} & 84.5 {\scriptsize($\downarrow$2.0)} & \textbf{1946} {\scriptsize($\downarrow$1159)} & 71.8 {\scriptsize($\downarrow$1.8)} & \textbf{3346} {\scriptsize($\downarrow$2141)} & 37.9 {\scriptsize($\downarrow$1.0)} & \textbf{2188} {\scriptsize($\downarrow$2914)} & 48.1 {\scriptsize($\downarrow$1.9)} & \textbf{3323} {\scriptsize($\downarrow$2324)} \\
+ \textbf{VSRM-PPO (ours)} & \underline{39.2} {\scriptsize($\downarrow$1.4)} & 6744 {\scriptsize($\downarrow$1771)} & \textbf{31.3} {\scriptsize($\uparrow$0.0)} & 6133 {\scriptsize($\downarrow$1896)} & \underline{85.5} {\scriptsize($\downarrow$1.0)} & 2391 {\scriptsize($\downarrow$714)} & \underline{73.5} {\scriptsize($\downarrow$0.1)} & \underline{4074} {\scriptsize($\downarrow$1413)} & \textbf{39.5} {\scriptsize($\uparrow$0.6)} & \underline{2231} {\scriptsize($\downarrow$2871)} & \textbf{50.1} {\scriptsize($\uparrow$0.1)} & 4414 {\scriptsize($\downarrow$1233)} \\
+ \textbf{VSRM-R++ (ours)} & \textbf{39.3} {\scriptsize($\downarrow$1.4)} & \underline{6668} {\scriptsize($\downarrow$1847)} & \underline{30.9} {\scriptsize($\downarrow$0.4)} & \underline{6103} {\scriptsize($\downarrow$1926)} & \textbf{85.8} {\scriptsize($\downarrow$0.7)} & \underline{2275} {\scriptsize($\downarrow$830)} & \textbf{74.7} {\scriptsize($\uparrow$1.1)} & 4082 {\scriptsize($\downarrow$1405)} & \underline{38.4} {\scriptsize($\downarrow$0.5)} & 2319 {\scriptsize($\downarrow$2783)} & \underline{49.4} {\scriptsize($\downarrow$0.6)} & \underline{4389} {\scriptsize($\downarrow$1258)} \\
\bottomrule
\end{tabular}%
}
\caption{Extensive experiments are conducted on several widely used benchmarks across multiple mathematical domains, comparing our method with several related approaches. Here, VSRM-PPO refers to training the base model by integrating our proposed VSRM with the PPO algorithm, while VSRM-R++ denotes the combination of VSRM with the Reinforce++ algorithm.}
\label{tab:main}
\end{table*}

Table \ref{tab:main} presents a comparison of VSRM with several related works across six benchmarks, organized into three sections according to the base model used: DS-Distill-1.5B, DS-Distill-7B, and DeepScaleR. The compared methods include LC-R1\cite{cheng2025optimizinglengthcompressionlarge}, ThinkPrune\cite{hou2025thinkprunepruninglongchainofthought}, AdaptThink\cite{zhang2025adaptthinkreasoningmodelslearn}, Efficient-Reasoning\cite{arora2025traininglanguagemodelsreason}, and GPRO-LEAD\cite{zhang2025grpoleaddifficultyawarereinforcementlearning}.

For experiments with DS-Distill-1.5B as the base model, our method consistently achieve the best or second-best performance across all benchmarks, while largely preserving or even slightly improving the original model’s capabilities on tasks such as AIME24, AIME25, AMC, and Minerva. Efficient Reasoning produce results comparable to ours on several benchmarks and achieve the best score on OlympiadBench; however, its output length was much longer, indicating a lack of efficiency. AdaptThink achieve slightly better output length compression than our method on MATH-500, AMC23, and Minerva, but this high level of conciseness restrict the model’s ability to explore different solution strategies. As a result, AdaptThink struggle in terms of overall performance and consistently lag behind our method. As for the other methods, both their performance and efficiency are generally less competitive compared to VSRM.

A similar trend is observed when using DS-Distill-7B as the base model. Both GRPO-LEAD and Efficient Reasoning adopt a strategy of sacrificing less compression to achieve better performance, resulting in performance levels comparable to ours. However, their efficiency is much lower than that of VSRM, with only minimal reductions in token count—the smallest among all evaluated methods. These two sets of experiments together indicate that, on distilled models, our method not only alleviates the overthinking problem but also has the potential to slightly improve overall model performance. Other methods also fall short of our approach when considering both performance and efficiency in a comprehensive manner.


DeepScaleR has already undergone systematic and thorough reinforcement learning, and its performance is nearly saturated. Consequently, further fine-tuning on this model often results in performance degradation, which explains why few related works use DeepScaleR as a base model. However, in practical model training pipelines, such as for Qwen3\cite{yang2025qwen3technicalreport}, training is typically conducted in multiple stages. Thus, efficiency optimization must ultimately target already strong LRMs that have been reinforced. In this work, we use DeepScaleR as the base model for our experiments. Results show that VSRM compresses output length across all benchmarks while maintaining performance close to the original DeepScaleR. By contrast, although ThinkPrune-2k achieves greater output length reduction, it compromises DeepScaleR’s strong reasoning ability, which runs counter to the original motivation of addressing the overthinking problem.

Across all experimental results, our method consistently reduces redundant output and mitigates the overthinking problem for models of various sizes and types, while preserving the original reasoning capability. Among all compared methods, VSRM achieves an excellent balance between performance and efficiency, demonstrating its superiority in fundamentally addressing the overthinking issue.

\subsection{Ablation Study}

\begin{table*}[t]
\centering
\small
\resizebox{\textwidth}{!}{%
\renewcommand{\arraystretch}{1.2}
\begin{tabular}{lcccccccccccc}
\toprule
\multirow{2.5}{*}{\textbf{Method}} & \multicolumn{2}{c}{\textbf{AIME24}} & \multicolumn{2}{c}{\textbf{AIME25}} & \multicolumn{2}{c}{\textbf{MATH-500}} & \multicolumn{2}{c}{\textbf{AMC23}} & \multicolumn{2}{c}{\textbf{Minerva}} & \multicolumn{2}{c}{\textbf{OlympiadBench}} \\

\cmidrule(lr){2-3} \cmidrule(lr){4-5} \cmidrule(lr){6-7} \cmidrule(lr){8-9} \cmidrule(lr){10-11} \cmidrule(l){12-13}

& pass@1$\uparrow$ & Length$\downarrow$ & pass@1$\uparrow$ & Length$\downarrow$ & pass@1$\uparrow$ & Length$\downarrow$ & pass@1$\uparrow$ & Length$\downarrow$ & pass@1$\uparrow$ & Length$\downarrow$ & pass@1$\uparrow$ & Length$\downarrow$ \\
\midrule
\textbf{DS-Distill-1.5B} & 28.8 & 12605 & 22.8 & 12444 & 82.4 & 4960 & 62.9 & 8577 & 34.2 & 6929 & 43.1 & 9249 \\
Original PPO & 29.0 {\scriptsize($\uparrow$0.2)} & 11111 {\scriptsize($\downarrow$1494)} & 22.8 {\scriptsize($\uparrow$0.0)} & 10545 {\scriptsize($\downarrow$1899)} & 82.3 {\scriptsize($\downarrow$0.1)} & 3968 {\scriptsize($\downarrow$992)} & 63.9 {\scriptsize($\uparrow$1.0)} & 7294 {\scriptsize($\downarrow$1283)} & 34.4 {\scriptsize($\uparrow$0.2)} & 5205 {\scriptsize($\downarrow$1724)} & 43.4 {\scriptsize($\uparrow$0.3)} & 7927 {\scriptsize($\downarrow$1322)} \\
VSRM-PPO w/ penalty & 28.6 {\scriptsize($\downarrow$0.2)} & 7533 {\scriptsize($\downarrow$5072)} & 20.9 {\scriptsize($\downarrow$1.9)} & 7369 {\scriptsize($\downarrow$5075)} & 80.9 {\scriptsize($\downarrow$1.5)} & 2615 {\scriptsize($\downarrow$2345)} & 64.1 {\scriptsize($\uparrow$1.2)} & 4695 {\scriptsize($\downarrow$3882)} & 34.2 {\scriptsize($\uparrow$0.0)} & 2712 {\scriptsize($\downarrow$4217)} & 42.9 {\scriptsize($\downarrow$0.2)} & 5102 {\scriptsize($\downarrow$4147)} \\
VSRM-PPO w/o decay & 28.8 {\scriptsize($\uparrow$0.0)} & 8638 {\scriptsize($\downarrow$3967)} & 22.6 {\scriptsize($\downarrow$0.2)} & 8451 {\scriptsize($\downarrow$3993)} & 81.1 {\scriptsize($\downarrow$1.3)} & 2990 {\scriptsize($\downarrow$1970)} & 63.4 {\scriptsize($\uparrow$0.5)} & 5618 {\scriptsize($\downarrow$2959)} & 34.1 {\scriptsize($\downarrow$0.1)} & 3288 {\scriptsize($\downarrow$3641)} & 43.1 {\scriptsize($\uparrow$0.0)} & 5935 {\scriptsize($\downarrow$3314)} \\
\textbf{VSRM-PPO} & \textbf{29.2} {\scriptsize($\uparrow$0.4)} & \textbf{7065} {\scriptsize($\downarrow$5540)} & \textbf{23.0} {\scriptsize($\uparrow$0.2)} & \textbf{6621} {\scriptsize($\downarrow$5823)} & \textbf{82.2} {\scriptsize($\downarrow$0.2)} & \textbf{2400} {\scriptsize($\downarrow$2560)} & \textbf{64.9} {\scriptsize($\uparrow$2.0)} & \textbf{4153} {\scriptsize($\downarrow$4424)} & \textbf{34.4} {\scriptsize($\uparrow$0.2)} & \textbf{2273} {\scriptsize($\downarrow$4656)} & \textbf{43.2} {\scriptsize($\uparrow$0.1)} & \textbf{4499} {\scriptsize($\downarrow$4750)} \\
\bottomrule
\end{tabular}%
}
\caption{Ablation experiments for our proposed VSRM-PPO method. We investigate the impact of each component by incrementally adding them to the baseline model. The evaluation is conducted on the DS-Distill-1.5B model across six standard mathematical reasoning benchmarks.}
\label{tab:ablation_study}
\end{table*}

Using DS-Distill-1.5B as the base model, we design ablation experiments to verify the contribution of VSRM and its individual components. All experiments are trained for the same number of steps to ensure fairness, and the results are shown in Table 3. 

First, we perform standard PPO-based reinforcement learning to the base model. We observe some improvement in performance on certain benchmarks, along with a slight reduction in output length, likely because the format constraints of RL reward discourage excessively long outputs.

Next, to explore the impact of different outcome reward designs, we adopt the length-penalized reward function proposed by \citet{yeo2025demystifying}, instead of the simple binary outcome reward. The results show that explicit length penalties do not further compress output length as expected; in fact, the compression is less effective than with VSRM-PPO. This is likely because the more complex outcome reward interferes with the effectiveness of the process-level rewards, leading to suboptimal results. 

Finally, we evaluate the impact of the propagation decay mechanism in VSRM, which is designed to deliver denser reward signals and guide the model toward correct reasoning paths more efficiently. Results show that while simple differential rewards can serve as stepwise feedback, their ability to suppress output length and maintain performance is significantly limited without propagation decay. This highlights the role of propagation decay in providing finer-grained supervision, enabling the model to better distinguish between effective and ineffective steps.

These ablation experiments illustrate the relationships and roles of each component of our proposed method, and demonstrate the rationality and effectiveness of VSRM.

\subsection{Analysis on Reasoning Behavior}

\begin{table}[h!]
\centering
\footnotesize
\begin{tabular}{l|c}
\toprule
\textbf{Model} & \textbf{Detected Overthinking} \\
\midrule
VSRM-PPO(DeepScaleR) & 132/500 \\
VSRM-PPO(DS-Distill-1.5B) & 126/500 \\
\bottomrule
\end{tabular}
\caption{Detecting ineffective steps via DeepSeek-R1}
\label{tb3}
\end{table}

Through the above experimental analysis, the effectiveness of our proposed method has been thoroughly validated. To further investigate the impact of VSRM on reasoning patterns, we conduct an empirical analysis in this section.

First, we revisit the example analyzed in Figure \ref{fig:case}, where DeepScaleR repeatedly self-reflects on the question of how many negative integers are in $[-500,0)$ within its generated response, ultimately arriving at an incorrect conclusion. These ineffective steps consume a large number of tokens. We examine all outputs (16 in total) produced by the model after VSRM-PPO reinforcement fine-tuning for this problem. In all reasoning processes, as long as there are no errors in the first half, the model no longer engages in repeated self-reflection on this question and was able to quickly summarize the correct answer. Several representative responses are provided in the Appendix. Similarly, to obtain broader and more robust conclusions, we once again adopt R1 to detect the degree of overthinking in models trained with the proposed VSRM. As shown in Table \ref{tb3}, although some instances of overthinking are still detected, this phenomenon is significantly mitigated compared to their respective base models. Considering that MATH-500 also contains some more challenging problems (level 5), a certain degree of self-reflection is to be expected. Therefore, we conclude that VSRM effectively reduces the occurrence of ineffective steps, thereby decreasing output length and fundamentally alleviating the overthinking problem.

\begin{figure}
    \centering
    \includegraphics[width=0.8\linewidth]{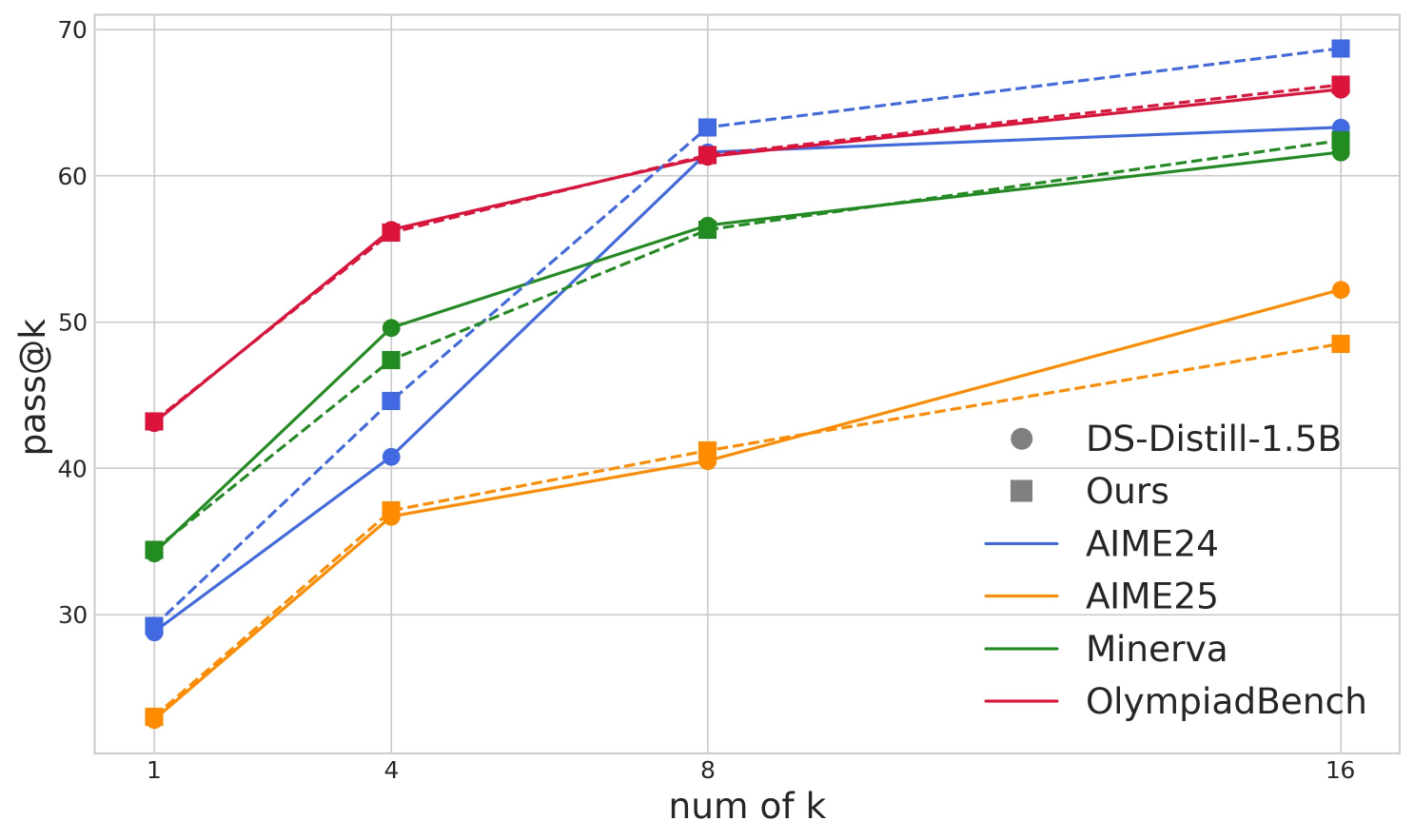}
    \caption{The increase in performance as the number of k grows reflects the model's diversity in exploring different problem-solving approaches.}
    \label{fig:passk}
\end{figure}

The above experiments demonstrate the effectiveness of VSRM in suppressing ineffective steps. Equally important, however, is its encouragement of effective steps, which we assess by evaluating the model’s exploration capability on challenging benchmarks.

To this end, we select four relatively difficult benchmarks—AIME24, AIME25, Minerva, and OlympiadBench—and compare DS-Distill-1.5B with its VSRM-PPO fine-tuned counterpart. We examine the change in the pass@k metric as k increases. The pass@k metric reflects the model’s ability to explore multiple approaches and strategies to eventually arrive at the correct answer. If the model’s exploration ability is restricted, pass@k will not show significant improvement as k increases; conversely, a steady increase indicates that the model’s exploration capacity is preserved. As shown in Figure \ref{fig:passk}, both the base model and VSRM-enhanced model exhibit a stable increase in pass@k as k grows. Moreover, the trend for our model is consistent with, and in some cases even surpasses, that of the base model. This demonstrates that VSRM does not hinder the model’s ability to explore valuable reasoning paths; on the contrary, it encourages the exploration of effective steps.

Together, these experiments provide strong evidence that VSRM not only effectively suppresses overthinking, but also preserves the model’s exploration ability, achieving output conciseness without sacrificing reasoning performance.

\section{Conclusion}

In this work, we focus on the overthinking problem of LRMs and revisit its essence. We propose to set “encouraging effective intermediate steps and penalizing ineffective ones” as the optimization objective. Based on this insight, we introduce a verifiable stepwise reward mechanism that assigns stepwise rewards to specific positions according to the performance of intermediate states along the reasoning trajectory. Extensive experiments conducted across multiple benchmarks on mathematical problems validate our insight. Our proposed method not only significantly mitigates the overthinking issue, but also preserves—and in some cases slightly enhances—the model's original reasoning capabilities, achieving efficient reasoning. We hope this work will inspire the community to advance large-scale efficient reasoning and facilitate the practical deployment of LRMs.

\bibliography{aaai2026}

\newpage
\appendix

\begin{figure*}[htbp]
    \centering
    \includegraphics[width=1.0\linewidth]{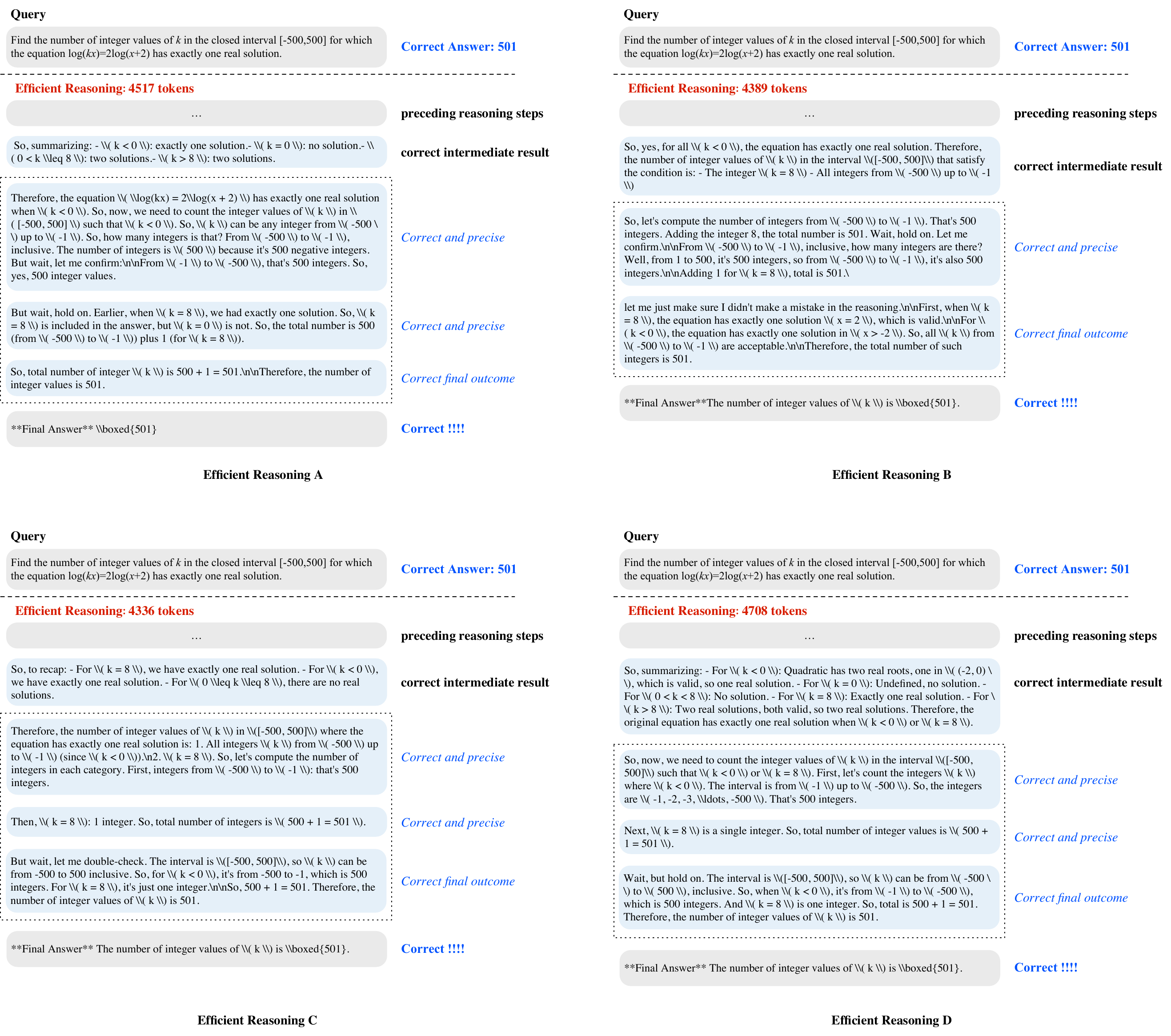}
    \caption{The model trained with VSRM-PPO effectively avoided generating a large number of ineffective steps. Although some reflection content still exists, appropriate reflection behavior can maintain performance to a certain extent.}
    \label{fig:case}
\end{figure*}

\section{More details about DS-R1 detecting ovethinking.}

Here we present the prompts used in Section \textbf{"Rethink the Essence of Overthinking Issue"} and \textbf{"Analysis on Reasoning Behavior"}.

\begin{tcolorbox}[breakable, colback=gray!10, colframe=gray!80, width=\linewidth ,title=Prompt used for DS-R1 to detect ovethinking.]

\textbf{\# Role}

You are a top-tier AI model behavior analyst. You specialize in analyzing the reasoning processes of models solving mathematical problems by comparing the \textbf{original problem} against the \textbf{generated solution}. Your mission is to identify whether a model's response is caught in a pattern of ``overthinking.''

\textbf{\# Background}

I am conducting an empirical study on the ``overthinking'' phenomenon in Large Language Models (LLMs) such as deepseek-R1. I have used these models to generate a batch of answers for problems from the math benchmarks. I now need you to analyze these ``problem-solution'' pairs to identify a specific type of inefficient reasoning.

\textbf{\# Core Task}

Please carefully review the provided \textbf{original math problem} and the model-generated \textbf{solution}. Your primary goal is to determine if it exhibits the following core issue:

\emph{``The model repeatedly struggles with a step that should be simple relative to the problem's complexity, remains unable to confirm its own answer, and therefore fails to advance to the next step in the reasoning process, leading to the consumption of an excessive number of tokens.''}

\textbf{\# Key Analysis Points}

In your analysis, please focus on the following aspects:

1. \textbf{Problem-Path Alignment}: Compare the problem's requirements with the model's solution path. Is the model's approach direct and efficient, or did it take unnecessary detours?
2. \textbf{Identify ``Sticking Points''}: Does the model get stuck on a specific concept or calculation that should be considered simple in the context of the overall problem's difficulty? (e.g., a basic algebraic manipulation, applying a common theorem, or a standard numerical calculation).
3. \textbf{Assess ``Struggling'' Behavior}: Does the model repeatedly propose multiple (even incorrect) approaches for the same point? Does it engage in constant self-correction and self-doubt without making tangible progress?
4. \textbf{Evaluate ``Efficiency''}: Compared to a standard or efficient method for solving this specific problem, did the model use a disproportionately large number of tokens or steps? Is there a significant amount of redundant text and calculation?
5. \textbf{Check for ``Progression Obstacles''}: Did this struggle clearly prevent the model from moving on to the next logical step of the solution?

\textbf{\# Analysis Report Structure}

Your analysis must include the following two sections:

1. \textbf{Detailed Analysis}:
    \begin{itemize}
        \item First, briefly summarize the core requirements of the problem.
        \item Next, summarize the model's overall problem-solving approach and comment on the reasonableness of its path.
        \item If you identify the ``overthinking'' phenomenon, you must pinpoint the exact step where it occurred. Quote the original text from the model's response as evidence and explain in detail how it demonstrates the characteristics of a ``sticking point,'' ``struggling,'' and ``inefficiency.''
        \item If you do not find this phenomenon, please state that the model's reasoning process was fluid and efficient.
    \end{itemize}

2. \textbf{Final Judgment}:
    \begin{itemize}
        \item At the very end of your analysis, you must state your conclusion in a single sentence, using one of the exact formats provided below.
    \end{itemize}

\textbf{\# Fixed Format for Final Judgment}

\texttt{[Final Judgment]: Yes}

or

\texttt{[Final Judgment]: No} (if the response does not exhibit clear signs of overthinking, and the reasoning process was relatively smooth)

---

\textbf{Please begin your analysis of the following problem, correct answer, and solution:}

The Original Problem: \{question\}

Correct answer: \{ground\_truth\}

The Model's Solution: \{prediction\}

\end{tcolorbox}

Although this has been clearly stated in the main text, to prevent any unnecessary misunderstanding, we clarify here once again: using prompts to instruct advanced LLMs to detect overthinking phenomena can only serve as a basic evaluation method, providing rough statistics on the occurrence of overthinking. Therefore, this approach cannot provide fine-grained, absolutely accurate and reliable evaluation, and cannot be used as a PRM.

\section{Detailed Experimental Setup}

This work is implemented based on the VeRL framework, with all relevant functionalities built upon it. The parameter settings are as follows. For basic reinforcement learning, we set the training batch size to 512, the maximum prompt length to 768, and the maximum response length to 8192. The actor’s PPO mini-batch size is 256, with a PPO micro-batch size per GPU of 8, and the rollout log probability micro-batch size per GPU is 16. The tensor model parallel size is set to 1, and the reference log probability micro-batch size per GPU is 16. For the critic, the PPO micro-batch size per GPU is 16 (note that the critic is not required for Reinforce++), and the total number of training epochs is 3. Regarding the proposed method, we set $n_\text{sub} = 5$, meaning that five candidate answers are generated for each subrollout, and $\text{max\_tokens}_\text{sub} = 768$, which is the maximum length of the generated answer for each subrollout. The choice of these two hyperparameters is mainly based on considerations of training speed, since each main trajectory generally contains 10--20 sub-trajectories on average, and increasing $n_\text{sub}$ would significantly increase the training time. All experiments are conducted on 8 A100 GPUs, with each training step taking approximately 20 minutes; as training progresses, the time required for each step gradually decreases. For evaluation, we use code adapted from evalscope~\cite{evalscope_2024}, an open-source LLM evaluation package, with evaluation settings following those described in the main text. The hyperparameters involved in step division are set as $C_\text{min} = 200$ and $I_\text{min} = 100$. For stepwise reward, the relevant hyperparameters are a reward decay factor of $0.7$, $L_\text{max} = 4$ (the maximum lookahead step), and $\epsilon = 10^{-5}$ (the tolerance for float equality checks).

\section{Case Study of Reasoning Pattern}
Here, in Figure \ref{fig:case}, we present the responses of DS-Distill-1.5B after VSRM-PPO training to the examples shown the main text, as discussed in Section “Analysis on Reasoning Behavior.” Due to the considerable length of individual samples, we only display the segments where the overthinking phenomenon occurs, selecting four representative samples for illustration.

Through these sample cases, it can be observed that the trained model, when encountering same simple question (how many integers are there within $[-500,0)$), is able to effectively avoid repetitive self-reflection and quickly arrive at the correct final conclusion, as long as the preceding reasoning steps are correct. This phenomenon validates the analysis in the “Analysis on Reasoning Behavior” section of the main text, demonstrating the effectiveness of VSRM in suppressing unnecessary reasoning steps.

\newpage
\section{Pseudocode of the Algorithm Involved in the Proposed Method}

\begin{algorithm}[h]
    \caption{Special Tokens Based Step Seperation}
    \label{Special Tokens Based Step Seperation}

    \KwIn{

        \quad$R = (t_0, t_1, \dots, t_N)$: A sequence of input tokens\\

        \quad$H_{p}$: A pattern to identify special tokens\\

        \quad$S_{p}$: A pattern to identify sentence-ending tokens\\

        \quad$C_{min}$: The minimum token offset\\

        \quad$I_{min}$: The minimum token interval between splits\\

    }

    \KwOut{
    
    \quad$P_{sorted}$: A sorted list of splitting token indices
    
    }
    
    \BlankLine

    \SetKwFunction{FStepSep}{StepSep}
    \SetKwProg{Fn}{Function}{:}{}
    \Fn{\FStepSep{$R, H_{\text{p}}, S_{\text{p}}, C_{\text{min}}, I_{\text{min}}$}}{



        $T \leftarrow \text{GetThinking}(R, \text{`\textless THINK\textgreater'}, \text{`\textless/THINK\textgreater'})$\;
        \If{$T$ is empty}{
            \Return $\emptyset$\;
            
        }

        $P \leftarrow \emptyset$\;
        
        $idx_{end} \leftarrow \text{EndIdx}(T, R)$\;
        
        $idx_{start} \leftarrow \text{StartIdx}(T, R)$\;

        \ForEach{$t_i$ in $\text{FindMatchedTokens}(H_{p},T[idx_{start},idx_{end}])$}{
        
            $idx_{token} \leftarrow \text{GlobalIndex}(t_i, R)$\;
            
            \If{($idx_{token} - idx_{start} < C_{min}$) \textbf{\upshape or} ($idx_{end} \neq -1$ \textbf{\upshape and} $idx_{token} - idx_{end} < I_{min}$)}{
            
                \textbf{continue}\;
            }
            $idx_{split} \leftarrow \text{FindPriorEnd}(R, idx_{token}, idx_{start}, S_{p})$\;

            \If{$idx_{split}$ \KwSty{is} null}{
                $idx_{split} \leftarrow idx_{token}$\; 
            }
            $P \leftarrow P \cup \{idx_{split}\}$\;
            $idx_{end} \leftarrow idx_{split}$\;
        }
        \KwRet \text{Sorted}($P$)\;
    }
\end{algorithm}

\begin{algorithm}[t]
    \caption{Stepwise Reward Computation}
    \label{alg:step_reward_computation_symbolic}

    \KwIn{

        \quad$A_{sub}$: A tensor of sub-rollout accuracies\\

        \quad$P_{token}$: A tensor of high-entropy token positions\\

        \quad$L_{valid}$: A tensor of valid response lengths\\

        \quad$\gamma$: The reward decay factor\\

        \quad$L_{max}$: The maximum lookahead steps\\

        \quad$\epsilon$: The tolerance for float equality checks\\

    }
    
    \KwOut{

        \quad$R_{step}$: A sparse tensor of step rewards\;
    }
    
    \BlankLine

    \SetKwFunction{csr}{ComputeStepReward}
    \SetKwProg{Fn}{Function}{:}{}
    \Fn{\csr{$A_{sub}, P_{token}, L_{valid}, \gamma, L_{max}, \epsilon$}}{
    
        $R_{step} \leftarrow \text{ZerosLike}(P_{token})$\;
        
        \For{\text{each sample} $i$ \text{in batch}}{
            $p_{list}, a_{list} \leftarrow P_{token}[i], A_{sub}[i]$\;
            
            \For{$j$, $idx_{token}$ in \text{Enumerate}($p_{list}$)}{
                $\delta \leftarrow a_{list}[j+1] - a_{list}[j]$\;
                
                \uIf{$\text{Abs}(\delta) < \epsilon$ }{
                    $r \leftarrow \text{RewardPropagationDecay}(j, a_{list}, \gamma, L_{max}, \epsilon)$\;
                }
                \Else{
                    $r \leftarrow \delta$\;
                }
                
                \If{$idx_{token}$ \KwSty{is} valid}{
                    $R_{step}[i, idx_{token}] \leftarrow r$\;
                }
            }
        }
        \KwRet $R_{step}$\;
    }
    
    \BlankLine


    \SetKwFunction{SubFn}{RewardPropagationDecay}
    \SetKwProg{Fn}{Function}{:}{}
    \Fn{\SubFn{$j, a_{list}, \gamma, L_{max}, \epsilon$}}{
        
        \uIf{\text{no such} $k$ \text{is found}}{
            \KwRet $0$\;
        }
        \Else{

            $\delta_{future} \leftarrow a_{list}[k+1] - a_{list}[k]$\;

            $dist \leftarrow k - j$\;

            \KwRet $\text{Sign}(\delta_{future}) \times |\delta_{future}| \times \gamma^{dist}$\;
        }
    }
\end{algorithm}

\end{document}